\documentclass[runningheads]{llncs}
\usepackage{graphicx}
\usepackage{enumerate}
\usepackage{dirtree}
\usepackage{enumitem}

\usepackage[encapsulated]{CJK}
\usepackage{graphicx}
\usepackage{amsfonts}
\usepackage{times}
\usepackage{soul}
\usepackage{url}
\usepackage[hidelinks]{hyperref}
\usepackage[utf8]{inputenc}
\usepackage{amsmath}
\usepackage{booktabs}
\usepackage{array}
\usepackage{amssymb}

\usepackage{xcolor}
\usepackage{tcolorbox}
\usepackage[linesnumbered,ruled]{algorithm2e}

\usepackage{arydshln}
\usepackage{multirow}
\usepackage{bm}
\usepackage{cases}
\usepackage{bbding}

\usepackage{CJKutf8}
\usepackage{color}

\definecolor{bluei}{RGB}{152,212,234}
\definecolor{redi}{RGB}{248,192,187}
\definecolor{greeni}{RGB}{111,139,62}
\definecolor{orange}{RGB}{247, 203, 172}

\newtcbox{\mybox}[1][red]
  {on line, arc = 1pt, outer arc = 2pt,
    colback = #1!10!white, colframe = #1!50!black,
    boxsep = 0pt, left = 1pt, right = 1pt, top = 1pt, bottom = 1pt,
    boxrule = 0pt, bottomrule = 0pt, toprule = 0pt}

\newcolumntype{L}[1]{>{\raggedright\let\newline\\\arraybackslash\hspace{0pt}}m{#1}}
\newcolumntype{C}[1]{>{\centering\let\newline\\\arraybackslash\hspace{0pt}}m{#1}}
\newcolumntype{R}[1]{>{\raggedleft\let\newline\\\arraybackslash\hspace{0pt}}m{#1}}

\begin{document}

\title{Nominal Compound Chain Extraction: A New Task for Semantic-enriched Lexical Chain}

\author{Bobo Li\inst{1} \and
Hao Fei\inst{1} \and
Yafeng Ren\inst{2} \and
Donghong Ji\inst{1}\thanks{Corresponding author.}
}

\institute{Key Laboratory of Aerospace Information Security and Trusted
Computing, Ministry of Education, School of Cyber Science and
Engineering, Wuhan University,Wuhan, China 
\and
Guangdong University of Foreign Studies, Guangzhou, China
\email{\{boboli,hao.fei,renyafeng,dhji\}@whu.edu.cn} 
}
\maketitle

\begin{abstract}
Lexical chain consists of cohesion words in a document, which implies underlying structure of a text, and thus facilitates downstream NLP tasks.
Nevertheless, existing work focuses on detecting the simple surface lexicons with shallow syntax associations, ignoring the semantic-aware lexical compounds as well as the latent semantic frames, (e.g., topic), which can be much more crucial for real-world NLP applications.
In this paper, we introduce a novel task, \textbf{N}ominal \textbf{C}ompound \textbf{C}hain \textbf{E}xtraction (NCCE), extracting and clustering all the nominal compounds that share identical semantic topics.
In addition, we model the task as a two-stage prediction (i.e., compound extraction and chain detection), which is handled via a proposed joint framework.
The model employs the BERT encoder to yield contextualized document representation.
Also HowNet is exploited as external resource for offering rich sememe information.
The experiments are based on our manually annotated corpus, and the results prove the necessity of the NCCE task as well as the effectiveness of our joint approach.
\keywords{Natural language processing \and Lexical chain  \and Nominal compound chain \and Semantic-aware compounds clustering \and  HowNet
}
\end{abstract}

\section{Introduction}
Lexical chain extraction (LCE) is an important research topic in the natural language processing (NLP) community, aiming to group cohesion words in a document into one cluster~\cite{01DBLP:journals/coling/MorrisH91}.
Lexical chain implies the underlying structure about the texts and provides practical cues for facilitating downstream NLP tasks, e.g., 
text summarization~\cite{barzilay1999using,mallick2019extractive}, 
keyword extraction~\cite{DBLP:conf/cicling/Carthy04,ercan2007using}, 
headline generation~\cite{DBLP:conf/aaai/XuYL10,sun2015event}, etc.

\begin{figure}[!t]
\centering 
\includegraphics[width=1.0\columnwidth]{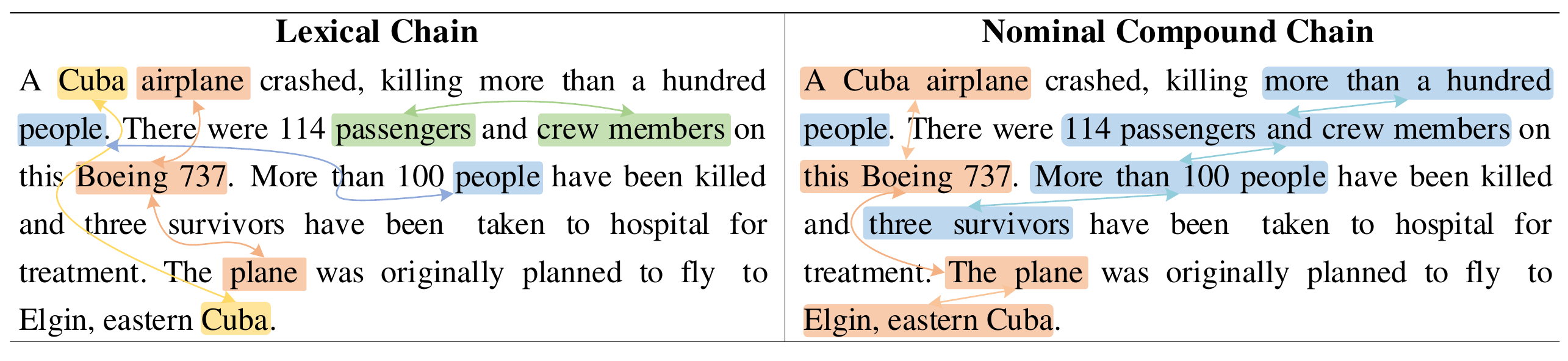}
\caption{
Lexical chain and nominal compound chain.
Mentions in same color indicate one chain.
}
\label{fig1}
\end{figure}

Fig. \ref{fig1} shows several lexical chain examples (under different colors).
Nevertheless, existing work about lexical chain extraction mainly focuses on detecting short lexicons under surface syntax associations, which inevitably leads to the information loss.
For instance, the lexical chain $\sharp_1$ (in \textcolor{orange}{$\blacksquare$}) [`\emph{airplane}'$\leftrightarrow$`\emph{Boeing 737}'$\leftrightarrow$`\emph{plane}'] entails the shallow-semantically identical entities, while the underlying frames carrying latent topic information are left unused.

To sufficiently exploit the richer semantic information into the lexical chain, in this paper, we propose a new task, named \textbf{N}ominal \textbf{C}ompound \textbf{C}hain \textbf{E}xtraction (NCCE).
Compared with LCE, NCCE aims to extract and cluster all possible entities that 1) are long nominal compounds, instead of short lexicons, and 2) describe same topic mentions in detail.
The differences between lexical chain and nominal compound chain can be seen Fig. \ref{fig1}.
For instance, as a counterpart of $\sharp_1$, the nominal compound chain $\sharp_2$ (in \textcolor{orange}{$\blacksquare$}) 
[`\emph{A Cuba airplane}'$\leftrightarrow$`\emph{this Boeing 737}'$\leftrightarrow$`\emph{The plane}'$\leftrightarrow$`\emph{Elgin$\dots$Cuba}'] 
involves longer nominal phrases which describes a latent topic (i.e., \textbf{airplane}) with much elaborated information, e.g., amount, nationality, etc.

On the other hand, directly modeling NCCE with the traditional extraction methods for lexical chain,
however, can be problematic.
Intuitively, NCCE is more semantic-dependent and context-sensitive.
Furthermore, NCCE faces with long-range text spans, and demands a stronger encoder.
To this end, we first treat NCCE as a two-step prediction task, i.e., \emph{Nominal Compound Extraction} and \emph{Chain Detection}, via a joint model.
We then consider two strategies for enhancing the ability of document representation.
As shown in the framework (Fig. \ref{figure_model}), we first employ the BERT language model~\cite{devlin2019bert,biofeibbaa110} to provide contextualized representation.
Besides, we exploit HowNet~\cite{dong2006hownet} as external resource for enriching the document from semantic perspective, which is encoded via a graph convolutional network (GCN,~\cite{FEI2020102311,kipf2016semi}).
Nominal compound extractor makes predictions based on the fused contextualized representation.
Chain detector then takes as input the recognized compound representations, detecting all the possible chains.

Due to the absence of the NCCE benchmark, we manually annotated 2,450 documents based on Chinese News texts. 
The experimental result on the dataset shows that the proposed joint model achieves 70.2\% F1 score in nominal compound detection, and 59.7\% F1 score in overall NCCE, outperforming the baseline systems, showing the effectiveness of the proposed model for the task.
In addition, we apply extracted nominal compound chain to enrich the \emph{sentence ordering task}, verifying 
that the helps in document understanding from the nominal compound chains are greater than that from lexical chains.
We publish codes and data at \url{https://github.com/unikcc/NCCE}.

\section{Nominal Compound Chain Extraction}
In this section, we first elaborate the criteria for building nominal compound chain. 
We then make description on constructing the dataset for the NCCE task.

\noindent \emph{\textbf{Task definition.}}
Unlike the lexical chain where the basic units are words or phrases, nominal compounds can be considered as the longest noun phrases (NP) in sentences\footnote{Valid nominal compounds shouldn't contain particle word, i.e., \begin{CJK}{UTF8}{gbsn}`的'\end{CJK} in Chinese}, as illustrated in Fig. \ref{fig1}. 
Furthermore, the nominal compounds satisfying these conditions will be categorized into one common cluster, forming a nominal compound chain.
\begin{enumerate}[itemindent=1em,label=\alph*)]
    \item They recur without identity of co-reference. For example, She likes \textbf{\em apples}; She bought some \textbf{\em apples}.
    \item They refer to a semantically-identical entity. For example, \textbf{\em A Cuba airplane} crashed $\dots$; \textbf{\em The plane} was originally $\dots$.
    \item They belong to the same parent collection. For example, \textbf{\em More than 100 people} have been killed and \textbf{\em three survivors} have been taken to hospital for treatment.
    \item They follows a \emph{PART-WHOLE} relationship. For example, There were \textbf{\em 114 passengers and crew members} on this Boeing 737; \textbf{\em More than 100 people} have $\dots$.
    \item They share common elements as modification or core word. For example, \textbf{\em A Cuba airplane} crashed $\dots$; $\dots$ fly to \textbf{\em Elgin, eastern Cuba}.
\end{enumerate}

\paragraph{\textbf{Data construction.} }
We manually annotated a high-quality Chinese dataset for facilitating the task.
Specifically, the data is built upon Chinese news corpus\footnote{Source: \url{http://people.com.cn}, \url{http://xinhuanet.com}, \url{https://sohu.com/}, etc.}.
The dataset is annotated based on crowdsource, and then is proofread by language experts in Chinese, by which we can ensure the high consistency on labels, and guarantee the data quality.
The final data contains 2,450 documents and 26,760 nominal compounds for a total of 5,096 chains.
We randomly split the total data into training, development and test sets with 2,050, 200, 200 documents, respectively. 
Table \ref{table_data} shows the statistics of the dataset.

\begin{minipage}[!t]{0.97\textwidth}
\begin{minipage}{0.43\textwidth}
\makeatletter\def\@captype{table}\makeatother
\caption{\label{citation-guide_3}
The statistics of the dataset. }
\resizebox{1.0\columnwidth}{!}{
  \begin{tabular}{lrrr}
\hline
 & Training & Development & Test \\
 \hline
Document & 2,050 & 200 & 200 \\
Compound & 22,565 & 2,124 & 2,071 \\
Chain & 4,277 & 402 & 417 \\
\hline
Max. chain size & 27 & 22 & 19 \\
\hdashline
Avg. compound length& 6.04 & 6.03 & 6.10 \\
Median. compound length & 4& 4 & 4 \\
Max. compound length  & 153 & 83 & 78 \\
\hline
\end{tabular}
    }
\label{table_data}
   \end{minipage}
\quad
\begin{minipage}{0.52\textwidth}
\centering
\makeatletter\def\@captype{figure}\makeatother
\includegraphics[width=1.0\columnwidth]{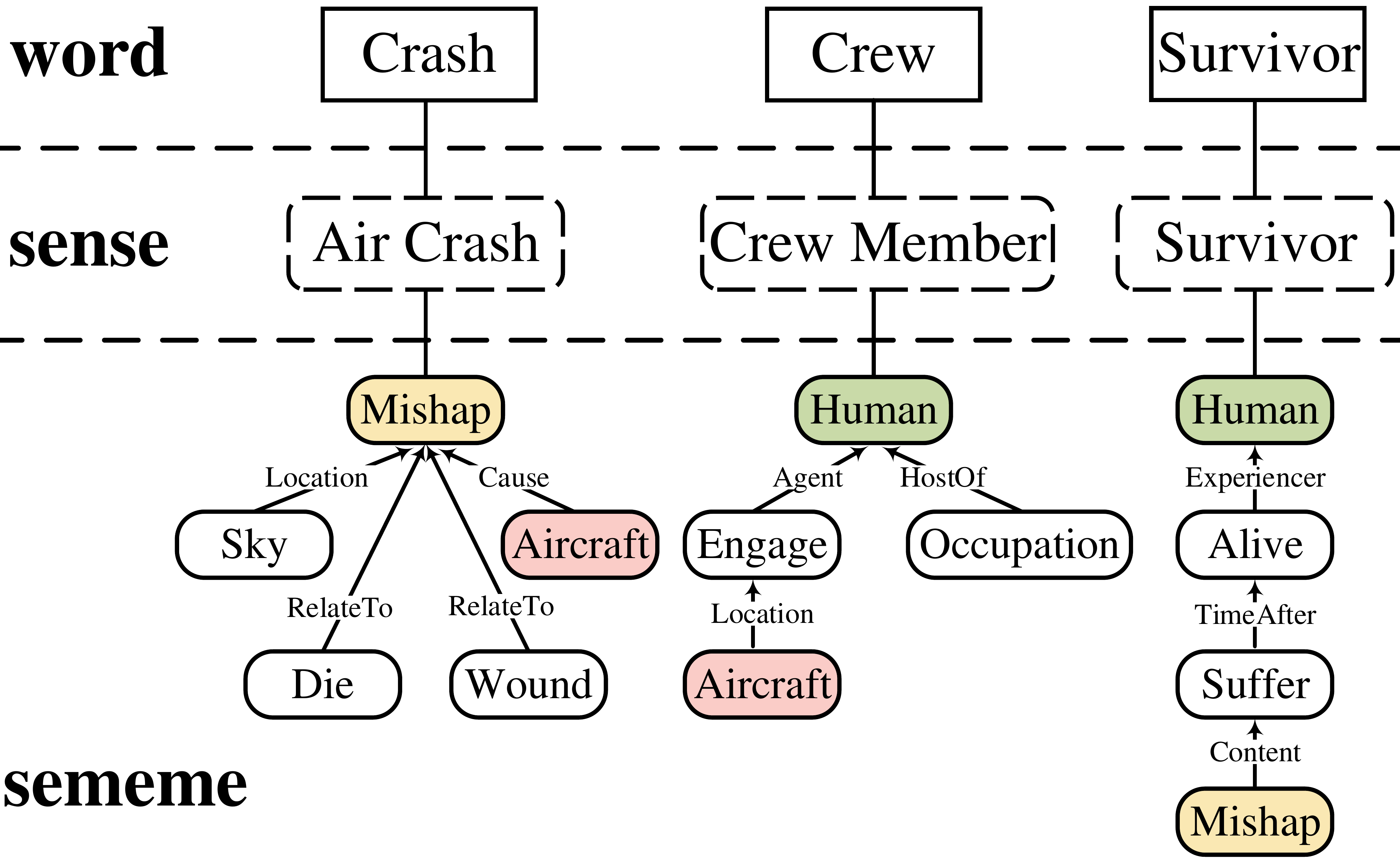}
\caption{
Word, sense and sememe in HowNet.
Same sememes share same color.}
\label{figure_hownet} 
\end{minipage}
\end{minipage}

\section{Preliminary}

\subsection{Task Modeling}
\label{task_section}

We model NCCE as a two-step prediction problem, i.e., nominal compound extraction and chain detection.
Given a document $D=\{w_1,\cdots,w_n\}$ ($w_i$ is a token), the final target of chain detection is to output all possible nominal compound chains $C=\{\widehat{T}_1,\cdots,\widehat{T}_p\}$.
The $i$-th compound chain $\widehat{T}_i$ is a certain subset of all candidate nominal compounds $T=\{c_1,\cdots,c_m\}$, where nominal compound $c_i=\{w_x,\cdots, w_y\}(1\le x\le y\le n)$ is detected in the first compound extraction stage\footnote{We assume that candidates and chains are not overlapped, and each candidate $c_i$ is assigned to a certain chain $\widehat{T}_{i'} (1\le i'\le p)$.}.

\paragraph{\textbf{Nominal compound extraction.}}
The first step of extracting nominal compound, is treated as a standard sequence labeling problem with the \emph{BIO} tagging scheme.
Here, `\emph{B}', `\emph{I}' and `\emph{O}' denote the beginning, inside and outside of the compound span, respectively. 
The extraction model will output the corresponding tags $\{t_1,t_2, \cdots, t_n\}$ ($t_i \in \{ B,I,O \}$) for each token $w_i$ in $D$, yielding all the candidate compounds $T$.

\paragraph{\textbf{Chain detection.}}
\label{chain_detection_label}

The second stage is to detect all chains based upon all recognized nominal compounds $T$\footnote{Suppose the elements in $T$ are sorted in ascending order.}, grouping the candidate $c_i$ into different chains $\widehat{T}_i$.
Following Lee et al., (2017), we transform the task into the problem of co-reference resolution \cite{lee2017end}, assigning each $c_i$ to an ancestor $c_k(1\le k \le i-1)$.
When each compound finds its ancestor, we finally obtain a set of nominal compound chains $C=\{\widehat{T}_1,\cdots,\widehat{T}_p\}$.

\subsection{HowNet for Enriching Semantics}

HowNet describes words or phrases via sememes, the minimum units for a semantic concepts~\cite{dong2006hownet}.
Recent studies show that HowNet can provide the rich acknowledged semantic representation and 
thus facilitate downstream NLP tasks \cite{DBLP:conf/acl/LiDLZS19,niu2017improved}.
Taking Fig. \ref{figure_hownet} as example, the word `\emph{survivor}' can be represented as the combination of sememes: `\emph{human}', `\emph{alive}', `\emph{suffer}' and `\emph{mishap}'. 
That is, the sememe set fully characterizes the semantic space of an entity and entails rich semantics.
Intuitively, such information can 
benefit both nominal compound detection and chain extraction.
For example, the `\emph{Survivor}' and `\emph{Crew}' share underlying sememe concepts `\emph{Human}', which can enhance the link between the nominal compounds `\emph{three survivors}' and `\emph{$\cdots$and crew members}' in Fig. \ref{fig1}, and ease the chain clustering.
Technically, we build a graph that connects different mentions by their senses and sememes for exchanging and enriching the semantics of input document.
Given $D$, each token $w_i$ connects to its sense and sememes, in which some of the sememes may link to other tokens $w_j$, which forms a graph architecture $\mathcal{G}=(\mathcal{V}_1,\mathcal{V}_2, \mathcal{E})$.
$\mathcal{V}_1$, $\mathcal{V}_2$ denotes the nodes including word tokens and senses/sememes, respectively, while $\mathcal{E}$ refers to the edges (in bi-directional) between node pairs.
In addition, we add self-loop on each node to maintain the enhanced representation \cite{liu2018jointly}.

\section{Joint NCCE Model}
\begin{figure}[!t]
\centering 
\includegraphics[width=0.92\columnwidth]{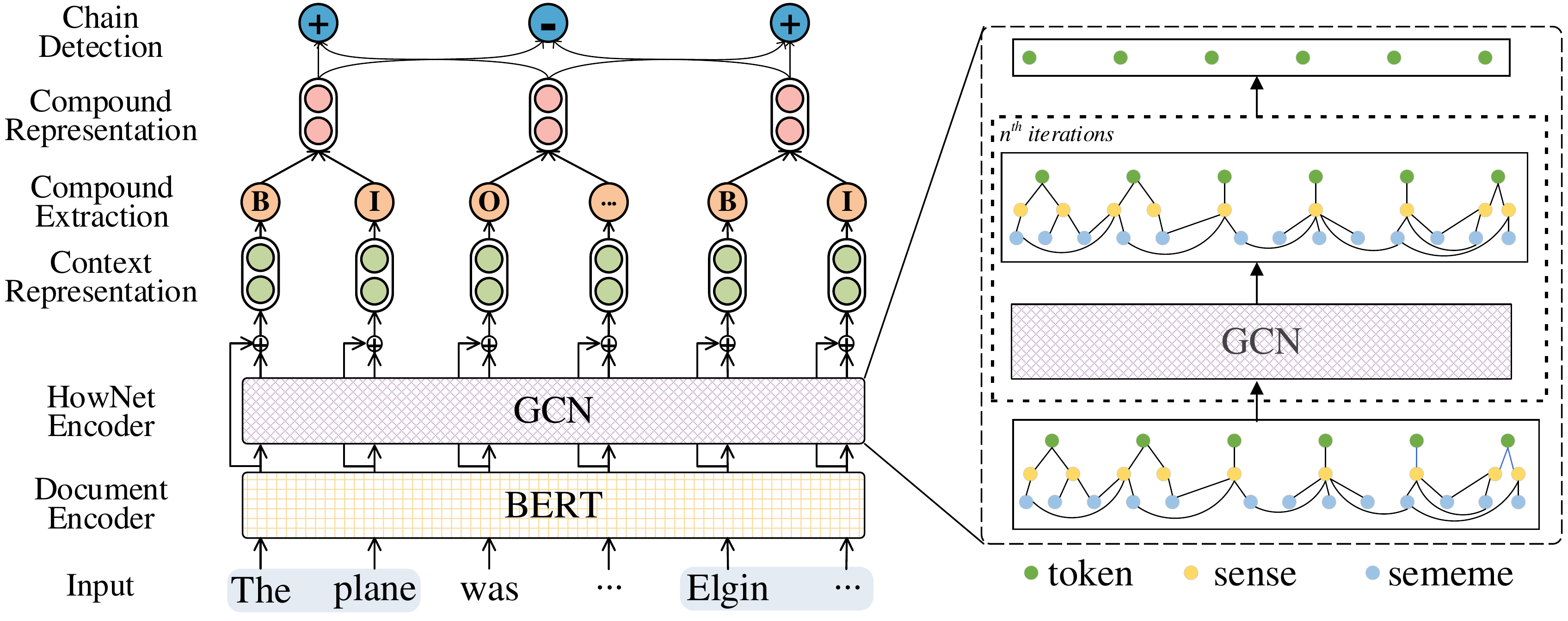}
\caption{
The architecture of the proposed model.
}
\label{figure_model}
\end{figure}

We further propose a joint neural model for NCCE.
As shown in Fig. \ref{figure_model}, the overall architecture mainly consists of four tiers.
First, the BERT encodes the document,
and the GCN model encodes HowNet for yielding the sememe representations.
Then, the document and sememe representations are concatenated and fed for nominal compounds detection via compound extractor.
Finally, based on the compound representations, the chain detector extracts all possible chain set.

\subsection{Document Encoder}

Unlike the sentence-level short texts, the documents are lengthy and entail more contexts.
Therefore, we employ the BERT language model as the document encoder.
BERT is built upon multi-head self-attention blocks.
Specifically, the attention weights are computed in parallel via:
\begin{equation}
\setlength\abovedisplayskip{2pt}
\setlength\belowdisplayskip{2pt}
    \bm{E} = \text{softmax}(\frac{\bm{Q} \bm{K}^{T}}{\sqrt{d}}) \bm{V} 
    = \text{softmax}( \frac{(t\cdot\bm{x}) \quad (t\cdot\bm{x})}{\sqrt{d}}) (t\cdot\bm{x}),
\end{equation}
where $Q$ (query), $K$ (key) and $V$ (value) are in practice the same input $\bm{x}$ scaled $t$ times.
BERT encoder yields contextualized representation $\bm{R}^{out}_i$ for each token $w_i$:
\begin{equation}
\label{bertEncoder}
\setlength\abovedisplayskip{2pt}
\setlength\belowdisplayskip{2pt}
\{ \bm{R}^{out}_{1},\cdots,\bm{R}^{out}_{n}\} = \text{BERT}( \{ \bm{R}^{in}_{1}, \cdots, \bm{R}^{in}_{n}\} ),
\end{equation}
where $\bm{R}^{out}_{i}$ is the output representation and $\bm{R}^{in}_{i}$ is the input representation.

\subsection{HowNet Encoder}

As we model HowNet as a graph $\mathcal{G}=(\mathcal{V}_1,\mathcal{V}_2,\mathcal{E})$, based on the input document, we use a GCN to learn the representation.
We first apply a linear transformation on token representation $h_i = \bm{W}^{t}\bm{R}_i^{out} + b^t$ for the initialization of GCN.
The sense or sememe nodes representations $e_i$ is obtained via a trainable embedding matrix $E \in \mathbb{R}^{m^{'} \times d}$.
Then we concatenate the representation of nodes in $\mathcal{V}_1$ and $\mathcal{V}_2$ for the GCN encoder:
\begin{equation}
\setlength\abovedisplayskip{2pt}
\setlength\belowdisplayskip{2pt}
\bm{N}^{0} = [h_1,h_2,\dots,h_n,e_1,e_2,\dots,e_m] \,.
\end{equation}
For the $(l+1)$-layer GCN, the update process can be denoted as:
\begin{equation}
\setlength\abovedisplayskip{2pt}
\setlength\belowdisplayskip{2pt}
\bm{N}^{l+1} = f(\tilde{\bm{D}}^{-\frac{1}{2}}\bm{A}\tilde{D}^{\frac{1}{2}}\bm{N}^{l}\bm{W}^{l}) \,,
\end{equation}
where $\bm{A}$ is adjacency matrix of the graph $\mathcal{G}$, $f$ is a activate function and $\tilde{\bm{D}}_{ii}=\sum_{j=1}\bm{A}_{ij}$ is a diagonal matrix. $\bm{W}^{l} \in \mathbb{R}^{d \times d}$ is the layer specific weight parameter. 
The updated token representation $\bm{v}_i$ will be used as the final representation:
\begin{equation}
\setlength\abovedisplayskip{2pt}
\setlength\belowdisplayskip{2pt}
\{\bm{v}_1,\cdots,\bm{v}_n \}= \bm{N}^{l}[0:n, :],
\end{equation}
where $\bm{N}^{l} \in \mathbb{R}^{(n+m) \times d}$ is the representation from the last layer of GCN.

\subsection{Compound Extractor} 
\label{detection_subsection}

We intend to concatenate the representations from BERT and GCN for avoiding information redundancy.
However, such concatenation on $\bm{h}_i$ and $\bm{v}_i$ is not flexible enough.
Instead, we introduce a fusion gate to better coordinate the representation sources:
\begin{equation}
\setlength\abovedisplayskip{2pt}
\setlength\belowdisplayskip{2pt}
\begin{aligned}
\label{gate} \bm{c}_i &= \sigma( \bm{W}^{f_1}h_i + \bm{W}^{f_2}v_i + b^{f}), \\
\bm{u}_i &= \bm{c}_i \odot \bm{h}_i + (1 - \bm{c}_i) \odot \bm{v}_i,
\end{aligned}
\end{equation}
where $\bm{W}^{f_1}, \bm{W}^{f_2} \in \mathbb{R}^{d \times d}$ are trainable parameters. 
The fusion gate controls the contribution proportions from document contexts and external enriched semantics.
Then, the fused representation $\bm{u}_i$ is used to predict the tag via softmax $o_i = \text{softmax}(\bm{u}_i)$.

\subsection{Chain Detector}

For the recognized compound $c_i$ which consists of token spans $\{w_x,w_{x+1},\cdots,w_y\}$, we can obtain its representation $\bm{p}_i$:
\begin{equation}
\setlength\abovedisplayskip{2pt}
\setlength\belowdisplayskip{2pt}
\bm{p}_i = [ \bm{u}_{x} ; \bm{u}_{y} ; \bm{c}_{att} ; \bm{c}_{wid}] \,,
\end{equation}
where $\bm{u}_{x}$ and $\bm{u}_{y}$ are boundary representations.
$\bm{c}_{wid}$ is the embedding vector for the span width.
$\bm{c}_{att}$ is the span attention representation over tokens in current compound:
\begin{equation}
\setlength\abovedisplayskip{2pt}
\setlength\belowdisplayskip{2pt}
\label{span attention representation}
\bm v_{t} = \text{tanh}( \bm{W} \bm{u}_{t})  \,,
\bm \alpha_{t} = \text{softmax}(\bm v_{t})  \,,
\bm{c}_{att} = \sum^{y}_{t=x} \bm \alpha_{t} \cdot \bm{u}_{t}  \,.
\end{equation}
We next assign each $c_i$ to an ancestor $c_k(1\le k \le i-1)$.
For each pair of compounds $c_i$ and $c_j$, we compute the probability $p(y_{ij}|o_i, o_j, \theta) \in [0,1]$ belonging to one chain:
\begin{equation}
\setlength\abovedisplayskip{2pt}
\setlength\belowdisplayskip{2pt}
\begin{aligned}
r_{ij} &= [o_i; o_j; o_i * o_j; o_i + o_j], \\
p(y_{ij}|o_i,o_j,\theta) &= \text{softmax}(\bm{W}^r r_{ij} + b^r) \,,
\end{aligned}
\end{equation}
where $r_{ij}$ is the interaction representation of two compounds.

\paragraph{\textbf{Heuristic searching.}}
Given the set of ancestor probabilities $P_i=\{p_{i0},p_{i1},\cdots,p_{ii-1}\}$ for each compound $c_i$, the final nominal compound chains can be decoded heuristically.
Specifically, considering the max element $p_{ik}$ in $P_i$ which means that the $k$-th candidate $c_k$ is the most likely ancestor of $c_i$,
if $p_{ik}>\lambda$ ($\lambda \in (0,1)$ is a tunable threshold), $c_k$ can be the ancestor of $c_i$ and $c_i$ is thus added into the same chain. 
Otherwise, if none of candidate is assigned to be the ancestor of $c_i$, we will build a new chain and set $c_i$ as the first element of newly created chain. 
Performing the above heuristic searching for all elements, we can finally obtain all possible nominal compound chains.

\subsection{Learning}

The targets of the learning in compound extractor and chain detector are to minimize the following losses, respectively:
\begin{equation}
\setlength\abovedisplayskip{2pt}
\setlength\belowdisplayskip{2pt}
\begin{aligned}
\mathcal{L}_l &= -\sum_{j=0}^{n}\sum^{k}_{i=0} p_j(\hat{o}_i)log(p_j(o_i)) \,, \\
\mathcal{L}_r &= -\frac{1}{R}\sum_{i=1}^{R}\sum_{j=0}^{2}  {\hat{p}_{ij}log(p_{ij})} \,.
\end{aligned}
\end{equation}
where $p(\hat{o}_i)$ is the ground-truth probability, k is the size of tag set and n is the number of tokens.
$R$ is the count of compound-pairs in a document, $\hat{p}_{ij}$ and $p_{ij}$ are the gold and predicted probability that $c_i$ and $c_j$ belong to one chain.
In our joint training, we optimize the final loss, $\mathcal{L} = \mathcal{L}_l + \mu \mathcal{L}_r$,
where $\mu$ is a coupling coefficient.

\section{Experiments}

\subsection{Settings} 
We use the pre-trained weights in \emph{BERT-base-chinese} version\footnote{\url{https://github.com/google-research/bert}} to initialize the BERT encoder, which has 12 layers with 768 dimensions of hidden state.
We use Adam as the optimizer with an initial learning rate of 1e-5 with warm-up rate of 0.1 and L2 weight decay of 0.01, training with early-stop strategy. 
The batch size is 1.
The maximum length of sentence is 128.
A dropout layer with 0.2 is used after the encoder and the fusion layer.
The factors $\lambda$ and $\mu$ are set as 0.5 and 0.4, according to our development experiments.
The sememes of words are obtained by an open API, OpenHowNet\footnote{\url{https://openhownet.thunlp.org/}}.
We also re-implement the joint model for co-reference resolution \cite{lee2017end,lee2018higher} as our strong baseline.
For the nominal compound detection, we adopt precision, recall and F1-score as metrics.
We use MUC, $\rm B_3$ and $\rm CEAF_{\phi_4}$ to evaluate the chain detection\footnote{The scores are evaluated by the standard scripts of CONLL12: \url{http://conll.cemantix.org/2012}.}.

\subsection{Main Results}
\label{subsection_overall}

Table \ref{table_main} shows the results of pipeline and joint methods under different setting. 
First of all, we can find that our proposed joint model consistently outperforms the pipeline counterpart under all settings.
In contrast to pipeline, the joint model achieves the improvements of 0.1\% F1 score (70.2-70.1) on compound extraction, and 1.6\% F1 score (59.3-57.7) on chain detection, respectively.
In addition, the improvements in the second chine detection stage are more significant than that in the first compound extraction stage.
The possible reason is that, the joint model can mitigate the error propagation from the first extraction step, and perform dynamic adjustment for chain detection in training. 
Most prominently, when BERT is unavailable, we can notice that the performance drastically drops, with roughly 20\% F1 score decrease, for both two sub-tasks in pipeline and joint schemes.
This can be explained by that the pre-trained contextualized representation in BERT can greatly enrich the information capacity of documents, relieving the polysemy problem to some extent.
Such observation is consistent with the recent findings of BERT applications
\cite{devlin2019bert,biofeibbaa110}

\begin{table*}[!t]
\caption{
Results on the NCCE task.
\emph{w/o BERT} denotes that replace the BERT encoder with Bi-LSTM, 
and \emph{w/o HowNet} denotes removing the HowNet resource by the GCN encoder.
\emph{w/o gate} indicates replacing the gate fusion mechanism (Eq. \ref{gate}) with direct concatenation.
}
\centering
\begin{tabular}{lccccccc}
\hline
& \multicolumn{3}{c}{\texttt{Compound Extraction}} & \multicolumn{4}{c}{\texttt{Chain Detection}}\\
\cmidrule(r){2-4}\cmidrule(r){5-8}
& Precision & Recall & F1 & MUC(F1) & $\rm{B_3}$(F1) & $\rm{CEAF_{\phi_4}}$(F1) &  Avg. (F1)\\
\hline
\multicolumn{8}{l}{\textbf{Pipeline} :}\\
\quad Ours & 68.8 & 71.4 & 70.1 & 60.5 & 51.2 & 61.4 & 57.7\\
\quad\quad  w/o BERT & \underline{47.6} & \underline{56.6} & \underline{51.7} & \underline{39.1} & \underline{31.3} & \underline{41.4} & \underline{37.3}\\
\quad\quad  w/o HowNet & 68.9 & 68.3 & 68.6 & 59.6 & 50.3 & 60.0 & 56.6\\
\hline
\multicolumn{8}{l}{\textbf{Joint} :}\\
\quad CoRef &- &- &- & 48.7 & 40.7 & 50.7 & 46.7 \\
\quad CoRef+BERT &- &- &- & 59.5 & 50.6& 59.7 & 56.6 \\
\hdashline
\quad Ours & 70.3 & 70.0 & \textbf{70.2} &  61.6 & 53.7 & 63.7 & \textbf{59.7} \\
\quad\quad  w/o BERT & \underline{45.6} & \underline{60.2} & \underline{51.9} & \underline{43.1} & \underline{33.8} & \underline{42.0} & \underline{39.6}\\
\quad\quad  w/o HowNet & 67.8 & 69.3 & 68.5 & 60.1 & 51.0 & 60.2 & 57.1\\
\quad\quad  w/o gate & 67.4 & 71.4 & 69.4 & 60.9 & 52.3 & 63.1 & 58.8\\
\hline
\end{tabular}

\label{table_main}
\end{table*}

We also see that if the HowNet module is removed, both the pipeline and joint methods can witness notable performance drops.
However, the influence from HowNet seems comparably weaker, compared with the BERT encoder.
In addition, the usefulness of the HowNet resource is more significant for chain detection, compared with the one for the nominal compound extraction.
For example, the gap is 2.6\% F1(59.7-57.1) for chain detection while the drop is 1.7\% F1(70.2-68.5) for compound extraction in the joint model.
This is partially due to the fact that, the enhanced sememes information can promote the interactions between different nominal mentions chain, being much informative for the chain clustering, which is consistent with our initial intuition introduced in $\S$4.2.
Furthermore, we compare our joint model with a strong baseline, CoRef, a joint model for co-reference resolution\footnote{Since the original CoRef model does not support pipeline scheme, failing to extracting the mentions standalone, and thus we only present the result of chain detection.}.
From the results we can learn that the CoRef model is much competitive, and with BERT, it reaches a close equivalent-level results to ours (without HowNet version for fair comparison), with 56.6\% F1 score.
Nevertheless, our model with the help of HowNet can outperform CoRef by 3.1\% F1 score on NCCE.
Also the gate mechanism (Eq. \ref{gate}) can bring positive effects for the results.

\subsection{Discussion}

\paragraph{\textbf{Influence of compound lengths.}}
One key challenges of NCCE lies in extracting longer nominal compounds, which is more tricky compared with the shorter lexical words in LCE.
Here we study the influence of compound lengths for nominal mention detection under differing settings, including joint/pipeline model with/without HowNet and with/without BERT, respectively.
Fig. \ref{fig_mention_length} illustrates the results of different nominal compound lengths.
First of all, the nominal compounds with lengths in 5-10 words increase the extraction difficulty the most, while the results will decrease when lengths are larger than 14.
In addition, we can find that with HowNet or BERT, both the pipeline and joint model can consistently better solve longer compounds, especially those with the length $\ge$ 13.
This is partially because the external sememes from HowNet can improve the understanding ability of the document context, facilitating the detection.
In particular, the improvements for those compounds with length $\ge$ 14 by BERT are more significant.

\begin{minipage}[!t]{0.97\textwidth}
\begin{minipage}{0.49\textwidth}
\makeatletter\def\@captype{figure}\makeatother
\includegraphics[width=0.96\textwidth,trim=0 0 0 0,clip]{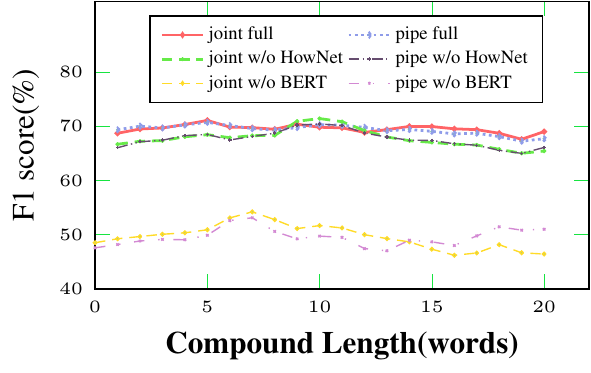}
\caption{F1 score against nominal mention length. The score is the average in a sliding window of size 7.
}
\label{fig_mention_length} 
\end{minipage}
\quad
\begin{minipage}{0.47\textwidth}
\makeatletter\def\@captype{figure}\makeatother
\includegraphics[width=0.96\textwidth,trim=0 0 0 0,clip]{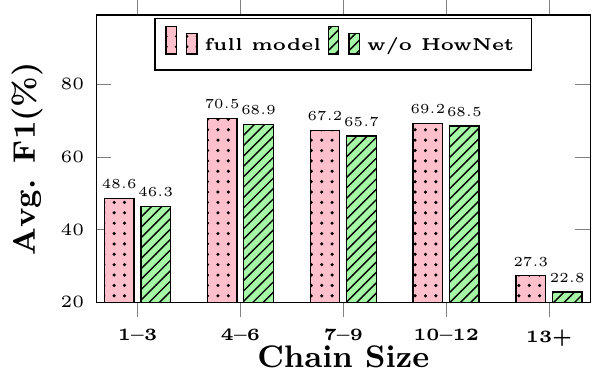}
\caption{
Performance of chain detection under varying chain size.
}
\label{figure_rank} 
\end{minipage}
\end{minipage}

\begin{table}[!t]
\caption{\label{citation-guide_4}
An example illustrating nominal compound chains (colored tokens) extracted by the models with/without HowNet resource, respectively.
}
\centering
\resizebox{1\columnwidth}{!}{
\begin{tabular}{p{8cm} c | c p{8cm}}
\hline
\hline
\multicolumn{1}{c}{ \begin{CJK}{UTF8}{gbsn}\fontsize{8.0pt}{\baselineskip}\selectfont $\bullet$ \textbf{Without HowNet}\end{CJK}} & \qquad\qquad  & \qquad\qquad  & \multicolumn{1}{c}{ \begin{CJK}{UTF8}{gbsn}\fontsize{8.0pt}{\baselineskip}\selectfont $\bullet$ \textbf{With HowNet}\end{CJK}} \\ 
\multicolumn{1}{m{7cm}}{ \begin{CJK}{UTF8}{gbsn}\fontsize{8.0pt}{\baselineskip}\selectfont \mybox[red]{高以翔躯体}被送回宝岛台湾...\mybox[red]{高以翔先生}的\mybox[red]{躯体}静卧于花丛中，身上的衣服干净肃穆。\end{CJK}}  & \qquad \qquad & \qquad & \multicolumn{1}{m{7cm}}{ \begin{CJK}{UTF8}{gbsn}\fontsize{8.0pt}{\baselineskip}\selectfont \mybox[blue]{高以翔躯体}被送回宝岛台湾...\mybox[blue]{高以翔先生}的\mybox[blue]{躯体}静卧于花丛中，\mybox[blue]{身上}的衣服干净肃穆。\end{CJK}}\\ 
\multicolumn{1}{m{7cm}}{ \begin{CJK}{UTF8}{gbsn}\fontsize{8.0pt}{\baselineskip}\selectfont \mybox[red]{Gao Yixiang}'s \mybox[red]{remains} was sent to Taiwan ... \mybox[red]{Sir Gao Yixiang}'s \mybox[red]{remains} was lying in the flowers quietly, and the clothes on the body were clean and solemn.\end{CJK}}  & \qquad\qquad  & \qquad\qquad  & \multicolumn{1}{m{7cm}}{  \begin{CJK}{UTF8}{gbsn}\fontsize{8.0pt}{\baselineskip}\selectfont \mybox[blue]{Gao Yixiang}'s \mybox[blue]{remains} was sent to Taiwan... \mybox[blue]{Sir Gao Yixiang}'s \mybox[blue]{remains} was lying in the flowerss quietly, and the clothes on \mybox[blue]{the body} were clean and solemn.\end{CJK}}\\ 
\hline
\multicolumn{4}{c}{\begin{CJK}{UTF8}{gbsn}\fontsize{8.0pt}{\baselineskip}\selectfont { $\bullet$ \textbf{Sememes from HowNet}}\end{CJK}}\\ 
\multicolumn{4}{l}{\begin{CJK}{UTF8}{gbsn}\fontsize{8.0pt}{\baselineskip}\selectfont \textbf{Word}: \emph{body}$|$身, \textbf{Sememes}:Def=\{part$|$部件:PartPosition=\{body$|$身\},domain=\{physiology$|$生理学\},whole=\{AnimalHuman$|$动物\}\}\end{CJK}}\\
\multicolumn{4}{l}{\begin{CJK}{UTF8}{gbsn}\fontsize{8.0pt}{\baselineskip}\selectfont \textbf{Word}: \emph{remain}$|$躯体, \textbf{Sememes}:Def=\{part$|$部件:PartPosition=\{body$|$身\},domain=\{physiology$|$生理学\},whole=\{AnimalHuman$|$动物\}\}\end{CJK}}\\
\hline
\hline
\end{tabular}
}

\label{table_case}
\end{table}

\paragraph{\textbf{Chain detection in varying chain size.}}

Chain size refers to the compound numbers within a chain, we further investigate the influences of HowNet for chain detection under different chain sizes.
As shown in Fig. \ref{figure_rank}, first, too shorter or too longer chains are more difficult to recognize, while the chains in [4,12] obtain better results.
We also see that the detection without HowNet is better, especially for those longer chains ($\ge$13).
The underlying reason can largely be that external sememe information from HowNet can provide more hints for the inference.

\paragraph{\textbf{Case study.}}
We explore how HowNet helps to facilitate the NCCE task.
Specifically, we empirically show an example extracted by our joint model, in Table \ref{table_case}, based on the test test. 
In particular, we clearly find that when the sememes from HowNet are employed, the extraction results become more complete.
For example, the surface compound words `\begin{CJK}{UTF8}{gbsn}{身上}\end{CJK}' (`the body') can be enriched by its sememes as listed in Table, which then will inspire the model to further yield correct extraction.
Without such links from words to external HowNet, the inference grows harder.

\subsection{Application}

\begin{table}[!t]
\caption{\label{citation-guide1}
Performance of the \texttt{sentence ordering} task under different input resources.
With /without 'type' indicates whether the syntactic role of the word in the sentence is considered when to build the graph.
}
\centering
\begin{tabular}{lccc}
\hline
\multicolumn{1}{l}{Input Resource} &   Accuracy  \qquad & PMR \qquad &  $\tau$ \\
\hline
\quad \multirow{1}{*}{Sentence}  &48.72 \qquad  & 21.00  \qquad &  66.57\\
\hline
\quad \multirow{1}{*}{Sentence+CW}  & 49.64 \qquad  & 19.49  \qquad & 66.62\\
\hline\hline
\multicolumn{4}{l}{\quad Sentence+LC}  \\
\quad \quad  w/o type & 49.84 \qquad  & 21.50  \qquad & 67.99 \\
\quad \quad with type  & 50.54  \qquad & 21.00  \qquad & 68.41 \\
\hline
\quad Sentence+NCC  & \textbf{51.87}  \qquad & \textbf{26.50}  \qquad &  \textbf{68.68} \\
\hline
\end{tabular}
\label{table_sentence_ordering}
\end{table}

As we emphasized earlier, compared with the traditional lexical chain, the nominal compound chain can be more expressive on rendering the underlying topics, providing details about the semantics of documents, which consequently can better facilitate the downstream NLP tasks.
To further quantify the usefulness of such characteristic, we here exploit the nominal compound chain extracted by our model for \texttt{sentence ordering}, a semantic-dependent task \cite{yin2019graph}.
Based on the state-of-the-art graph model in Yin et al., (2019)\cite{yin2019graph}, we first implement the task with raw sentence inputs, and besides we leverage the \emph{common words} (CW) as external resource\footnote{For more technical details, please refer to the raw paper of Yin et al., (2019)\cite{yin2019graph} }.
We then additionally extend the inputs with the lexical chains (LC) and nominal compound chains (NCC), respectively.
Technically, we utilize these external resources by building graphs, connecting the surface words with the corresponding nodes from the chains.
We follow the same metrics as Yin et al., (2019), including accuracy, PMR, and $\tau$.

Table \ref{table_sentence_ordering} shows the main results.
First, the comparison between top two rows indicates that the integration with enhanced resources can benefit the \texttt{sentence ordering} task.
We clearly find that the lexical chains help to give the improved task performance, than the common words.
Specifically, with more fine-grained information, the helpfulness is more evidence, which can be inferred from the results from \emph{with lexicon type} and \emph{without lexicon type}.
Most importantly, our proposed nominal compound chains can improve the result the most.
Significantly, the PMR metrics are increased by 5(26.50-21.50) compared with \emph{sentence+LC}, and 5.5(26.5-21.0) compared with the raw sentence input, respectively.
This shows the usefulness of our introduced nominal compound chain.

\section{Related Work}

Morris et al. (1991) pioneer the lexical cohesion (chain) task, a concept that arises from semantic relationships between words and provides a cue of text structures \cite{01DBLP:journals/coling/MorrisH91}.
Based in the concept of cohesive harmony, Remus et al. (2013) propose three knowledge-free methods to extract lexical chain from document automatically \cite{remus2013three}. 
Yunpeng et al. (2016) develop a method with semantic model for chain extraction, and prove that semantic features can be key signals to improve the task \cite{yunpeng2016using}.
Mascarell (2017) uses word embedding to compute the semantic similarity, and improves the results by considering the contextual information \cite{mascarell2017lexical}. 
Nevertheless, the lexical chain extraction involves in shallow lexicon knowledge, lacking the use of latent semantic information \cite{fei-etal-2020-cross,FeiZRJ20}, such as the topic information \cite{Fei9113297}, which limits the usefulness for the downstream tasks.
This motivates us to propose a novel task of Nominal Compound Chain Extraction.

Lexical chain extraction as one of the information extraction tasks \cite{FeiRJ20},  shares much technical similarities with the co-reference resolution task \cite{elango2005coreference}, as both of them model the task as a chain prediction problem. 
While the lexical chain extraction task focuses more on the semantic coherence between mentions, the latter aims to identify mentions of same entity, event or pronoun in groups.
Recently, an increasing number of neural co-reference resolution models have been proposed 
\cite{lee2017end,lee2018higher}, and greatly outperform previous machine learning models with hand-crafted features.
For example, Lee et al. (2017,2018) first propose end-to-end neural model rely on pairwise scoring of entity mentions.

\section{Conclusion}
In this work, we proposed a novel task, namely \emph{nominal compound chain extraction}, as the extension of the \emph{lexical chain extraction}.
The nominal compound chain can provide richer semantic information for rendering the underlying topics of documents.
To accomplish the extraction, we proposed a joint model, formulating the task as a two-step prediction problem, including \emph{Nominal Compound Extraction} and \emph{Chain Detection}.
We made use of the BERT contextualized language model, and enriched the semantics of input documents by leveraging the HowNet resource.
We manually annotated a dataset for the task, including 2,450 documents, 26,760 nominal compounds and 5,096 chains.
The experimental results showed that our proposed joint model gave better performance than the pipeline baselines and other joint models, offering a benchmark method for nominal compound chain extraction.
Further analysis indicated that both BERT and external HowNet resource can benefit the task, especially the BERT language model.

\section{Acknowledgments}

This work is supported by the National Natural Science Foundation of China (No.61772378, No.61702121), 
the National Key Research and Development Program of China (No.2017YFC1200500), 
the Humanities-Society Scientific Research Program of Ministry of Education (No.20YJA740062),
the Research Foundation of Ministry of Education of China (No.18JZD015), 
the Major Projects of the National Social Science Foundation of China (No.11\&ZD189),
the Key Project of State Language Commission of China (No.ZDI135-112) 
and Guangdong Basic and Applied Basic Research Foundation of China (No.2020A151501705).

\bibliographystyle{splncs04}
\bibliography{nlpcc2020}
\end{document}